\documentclass[conference]{IEEEtran}
\usepackage{times}
\usepackage{graphicx}
\usepackage{xcolor}
\usepackage{subfigure}
\usepackage{dsfont}
\usepackage{amsmath}
\usepackage[numbers]{natbib}
\usepackage{multicol}
\usepackage[bookmarks=true]{hyperref}

\title{Imitation Learning-Based Path Generation for the Complex Assembly of Deformable Objects}

\author{
\authorblockN{Yitaek Kim$^{1}$}

\authorblockA{SDU Robotics, \\
The Maersk Mc-Kinney Moller Institute,\\
        Syddansk Universitet, Campusvej 55, \\
        Odense M 5230, Denmark\\
Email: yik@mmmi.sdu.dk}
\and
\authorblockN{ Christoffer Sloth$^{\dagger}$}
\authorblockA{SDU Robotics, \\
The Maersk Mc-Kinney Moller Institute,\\
        Syddansk Universitet, Campusvej 55, \\
        Odense M 5230, Denmark\\
Email: chsl@mmmi.sdu.dk}
}
\begin{document}
\maketitle
\begin{abstract}
This paper investigates how learning can be used to ease the design of high-quality paths for the assembly of deformable objects. Object dynamics plays an important role when manipulating deformable objects; thus, detailed models are often used when conducting motion planning for deformable objects. We propose to use human demonstrations and learning to enable motion planning of deformable objects with only simple dynamical models of the objects. In particular, we use the offline collision-free path planning, to generate a large number of reference paths based on a simple model of the deformable object. Subsequently, we execute the collision-free paths on a robot with a compliant control such that a human can slightly modify the path to complete the task successfully. Finally, based on the virtual path data sets and the human corrected ones, we use behavior cloning (BC) to create a dexterous policy that follows one reference path to finish a given task.
\end{abstract}
\IEEEpeerreviewmaketitle
\section{Introduction}
Robotic assembly has been developed for many years and is an integral part of most factories. Nevertheless, the adoption of assembly of deformable objects has been slow compared with rigid objects because it requires an integrated system with complex modeling, perception, and manipulation. To ease the development of assembly for deformable objects, mathematical modeling, perception, and control of the objects have been studied \cite{RN1}. Furthermore, the manipulation of deformable objects often relies on a highly complicated reference path to finish a given task. To obtain the reference path, the approaches can be classified as detailed physics model-based and data-driven \cite{RN1}.

One example of the model-based approaches to defining a reference path is that \citet{RN2} models a small thin flexible object based on a viscoelastic joint model and then simulates a flipping of the object through sinusoidal trajectories as a reference path. There is also the way to use trajectory optimization with defining an objective function to obtain certain trajectories for folding a garment with physical properties such as friction forces \cite{RN3}. Similar to the previous one, another optimization approach is that CHOMP \cite{RN4} is used to optimize trajectories. However, although well-defined modeling of the objects is carried out, the model-based approaches have always a discrepancy between the physics model and the real world one due to unpredictable properties of the objects. The approach might also be impractical in a highly complex assembly process with deformable objects because it is too application-oriented to adapt to other industrial applications. 

A promising way to address this problem is data-driven methods by learning algorithms \cite{RN1}. Imitation learning and reinforcement learning have been used to obtain a nominal trajectory to achieve a given task \cite{imi-RN1}\cite{imi-RN2}\cite{imi-RN3}. To get the trajectory, the learning-based methods start with collecting human demonstrations. There are works to get reference motions for mounting a flexible tube \cite{RN7} and for knotting ropes \cite{RN8} through a teleoperation system. Also, combining robot execution with human demonstrations by collecting data from visual feedback shows high success rates in manipulating flexible ropes to be desired configurations \cite{RN9}. This approach can be easy to get reference paths compared to the model-based ones through intuitive human demonstrations, however, obtaining reference paths from visual feedback and teleoperation systems has disadvantages of not only that this way cannot be used in highly complex assembly tasks, but also it requires setting up additional equipment and lots of time-consuming demonstrations. Therefore, it is necessary to consider the approach to design a sophisticated reference path in terms of combining traditional path planning algorithms with learning-based methods.

In this paper, we present the pre-training method combining collision-free path planning, TrajOpt \cite{RN12} with Behavior Cloning (BC) \cite{RN13} to extract a high-quality reference path to finish a given task with a deformable object. By using TrajOpt, we define collision-free paths based on constraints and cost functions not only from the robot's CAD data but also from physics modeling of a rubber belt, Hunt-Crossley model \cite{RN14}, and we simultaneously obtain a large number of the paths via various waypoints. This means that it is not necessary to start the learning from scratch since we take advantage of already existent and traditional model-based control \cite{RN1}. To reduce the discrepancy between simulation and the real world, the reference paths or trajectories from the path planning algorithm are corrected by minimum human kinesthetic guides. We use position/force hybrid controller \cite{RN15} to be able to correct the paths by the guides while a robot follows along with the paths. To utilize data sets best, we generate new offline paths by fitting the original offline paths using the polynomial approach and combine them with human-guided paths to create virtual human-guided paths. This allows for a policy to learn with plenty of demonstration data. In order to imitate the data sets, BC \cite{RN13} is considered to create a policy network to produce a reference trajectory by using a neural network. We use Mean Square Error (MSE) \cite{RN13} as the objective function to mimic the reference paths \cite{RN17}. 

There are some advantages of using our approach. Firstly, since we consider the physics modeling of a deformable object, we can obtain rather reliable reference paths even though dealing with the non-linear component is not completely excellent. Secondly, after obtaining feasible reference paths, it is possible to extend new cost functions to optimize the paths in the learning algorithm in addition to imitating the paths, which can reduce the gap between simulation and the real world.

\section{Related Work}
\subsection{Motion Path Planning with deformable objects}
A Probabilistic Roadmap (PRM) is utilized to generate path planning with elastic objects in \cite{PRM-1} \cite{PRM-2}. However, the PRM approach requires high computational resources. On the other hand, optimization-based approaches like CHOMP \cite{RN4} and TrajOpt \cite{RN12} have the advantage of freely defining constraints for e.g. collision avoidance and physical constraints, in addition to cost functions for e.g. minimization of energy consumption or path length.
\subsection{Highly Complicated Task}
Some studies using human demonstrations address a rather complicated assembly such as insertion of an L-shaped part into another part \cite{RN11} and insertion of elastic ring-shaped objects into a cylinder \cite{RN10}. 
\subsection{Learning from Human Demonstrations}
Human demonstrations are used to represent how to execute a given task and offer skills for starting points to a robot. A single trajectory to tie knots in a flexible rope or flat towel is obtained through demonstrations \cite{RN24}. \citet{RN20} presents human demonstrations with a special glove onto which tactile sensors are embedded to learn how to grab an object.

\section{Method}
\subsection{Generation offline paths by TrajOpt }
TrajOpt \cite{RN12} uses sequential convex optimization to solve a non-convex optimization problem by iterating convex subproblems. TrajOpt is used for including constraints from the system dynamics and kinematic constraints to avoid collisions. The optimization problem is defined as: 

\begin{subequations}
\begin{IEEEeqnarray}{s,rCl'rCl'rCl}
minimize   & \IEEEeqnarraymulticol{9}{l}{f(\mathbf{x})} \\
subject to 
            & g_i(\mathbf{x}) &\le&0, & i &=& 1,2,\dots,n_{ineq} \\
            & h_i(\mathbf{x}) &=&0,  & i &=& 1,2,\dots,n_{eq} 
\end{IEEEeqnarray}
\end{subequations}
\noindent where $\mathbf{x}\in\mathds{R}^6$ is a vector of joint configurations. 

The following presents how TrajOpt is applied to the assembly of a rubber belt onto two pulleys. The objective of the optimization problem is the minimization of path length in joint space. To simplify the assembly of the rubber belt, tension is kept on the belt throughout the assembly \cite{RN22}. This is implemented by adding force constraints to the optimization problem as:
\begin{subequations}
\begin{IEEEeqnarray}{s,rCl'rCl'rCl}
    & g_1(\mathbf{x}) &=& F(\mathbf{x})-f_{upper}\\
    & g_2(\mathbf{x}) &=& f_{lower}-F(\mathbf{x})
\end{IEEEeqnarray}
\end{subequations}

\noindent where $f_{upper}$ and $f_{lower}$ are bounds on forces, and $F(\mathbf{x})$ is the belt force. The belt force is computed using a  Hunt-Crossley model \cite{RN14}. The model is defined as:
\begin{subequations}
\begin{IEEEeqnarray}{s,rCl'rCl'rCl}
    & F(\mathbf{x})&=&kx^{\beta}+\lambda{x^{\beta}\dot{x}}
\end{IEEEeqnarray}
\end{subequations}

\noindent where the displacement $x$ is obtained by robots' end-effector through calculating forward kinematics with $\mathbf{x\in\mathds{R}^6}$, and  $k$, $\beta$, $\lambda$ are the estimated parameters which are obtained by calculating non-linear least squares method, Levenberg–Marquardt algorithm \cite{RN23}. Finally, we generate offline reference paths through the path planning algorithm with constraints such as collision avoidance and keeping forces. The generated offline paths are defined as: 
\begin{subequations}
\begin{IEEEeqnarray}{s,rCl'rCl'rCl}
    & Path &=&\{p_{_0},p_{_1},p_{_2},\ldots,p_{_T}\} \label{subeq1} \\
    & p_{_T} &=&\{x,y,z,roll,pitch,yaw\}\nonumber
\end{IEEEeqnarray}
\end{subequations}
\noindent where $p_{_T}$ is the pose of the robot's end-effector at time $T$.

\subsection{Human Guided Paths}
Corrections from a human take place based on compliance control through a position/force hybrid controller when a robot moves along the given offline paths \eqref{subeq1}. Since there are reference paths, it is not necessary to demonstrate all paths by a human, and we only modify the paths to reduce the discrepancy between simulation and the real world. The human correction paths are specified as: 
\begin{subequations}
\begin{IEEEeqnarray}{s,rCl'rCl'rCl}
    & Path^* &=&\{p^{*}_{_0},p^{*}_{_1},p^{*}_{_2},\ldots,p^{*}_{_T}\} 
    \nonumber
\end{IEEEeqnarray}
\end{subequations}
\noindent where 
\begin{subequations}
\begin{IEEEeqnarray}{s,rCl'rCl'rCl}
    & p^{*}_{_T} &=&\{P^{*}_{_T},O^{*}_{_T}\},\nonumber\\
    & P^{*}_{_T} &=&\{x+\Delta{x},y+\Delta{y},z+\Delta{z}\}, \nonumber\\
    & O^{*}_{_T} &=&\{roll+\Delta{roll},pitch+\Delta{pitch},yaw+\Delta{yaw}\} \nonumber
\end{IEEEeqnarray}
\end{subequations}
$p^{*}_{_T}$ includes corrected position, $P^*_{_T}$ and orientation, $O^*_{_T}$.
\subsection{Generate Virtual Data Sets}
\begin{figure}[h!]
\centering{
{
\includegraphics[width=245pt , height=100pt]{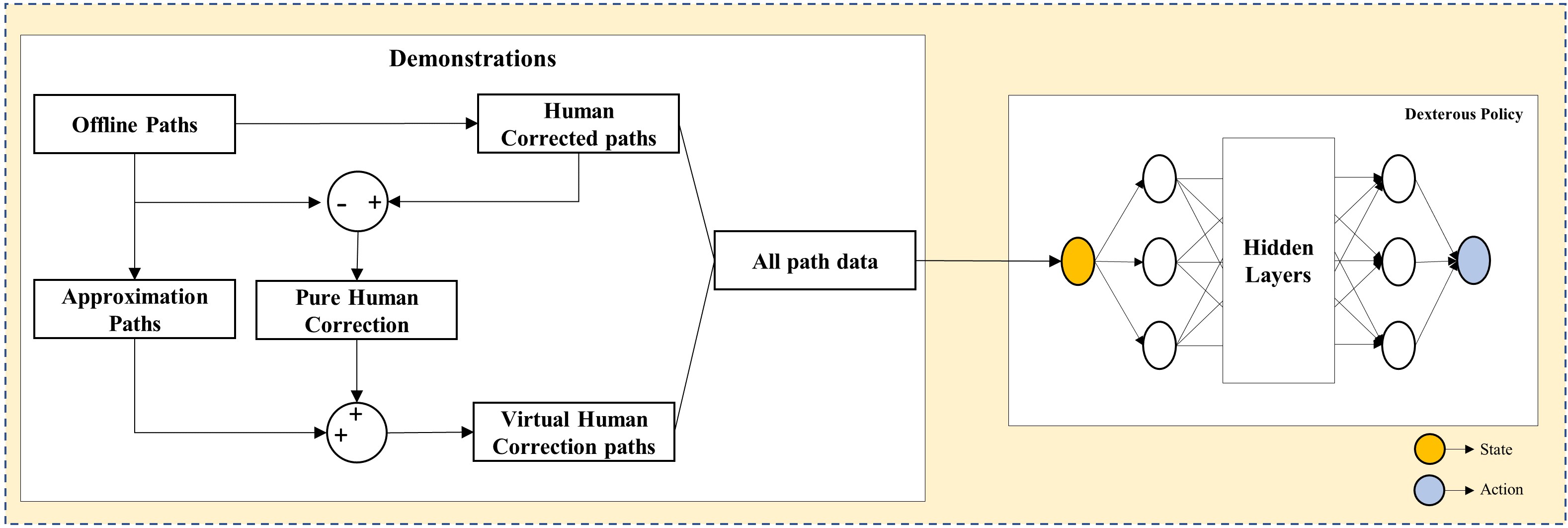}
\label{fig:outline}
}
\caption{
Overview of a combined approach for learning reference trajectory. 
}\label{fig:results}
}
\end{figure}

In order to obtain plenty of human demonstration data, we make virtual human correction data sets by using polynomial approximation. The overview of this is shown on the left side of Fig.~\ref{fig:results}. First of all, a previous offline path from the path planning algorithm is approximated in high order polynomial, which creates a new offline path independent of the previous one. Secondly, a pure human correction is obtained by subtracting the previous offline path from a human corrected path. And then we generate a virtual human corrected path by adding the approximated path to the pure human correction.

\subsection{Learn Reference Trajectory}
We use BC \cite{RN13} to imitate the demonstrations from simulation and a human. To get a policy to imitate the demonstrations, we firstly define states and actions as:
\begin{subequations}
\begin{IEEEeqnarray}{s,rCl'rCl'rCl}
    & S &=&\{\mathbf{s}_{_0},\mathbf{s}_{_1},\mathbf{s}_{_2},\ldots,\mathbf{s}_{_T}\},\nonumber\\
    & A &=&\{\mathbf{a}_{_0},\mathbf{a}_{_1},\mathbf{a}_{_2},\ldots,\mathbf{a}_{_T}\}\nonumber
\end{IEEEeqnarray}
\end{subequations}
\noindent where the states $s_{_T}$ are defined as a vector $\{x,y,z\}$, a position of the end-effector in relative to the center of a pulley to be assembled, and the actions $a_{_T}$ are defined the pose of end-effector in relative to robot's base frame. Based on the definition of the states and actions, we specify the demonstration data set (\ref{subeq2}) in order to generate the policy deciding which actions are suitable to a current state.
\begin{subequations}
\begin{IEEEeqnarray}{s,rCl'rCl'rCl}
    & \mathcal{D}_{set} &=&\{\mathcal{D}^{0},\mathcal{D}^{1},\ldots,\mathcal{D}^{N}\}\label{subeq2}
\end{IEEEeqnarray}
\end{subequations}
\noindent where $\mathcal{D}^{i} = \{\mathbf{s}_{_0},\mathbf{a}_{_0},\ldots,\mathbf{s}_{_T},\mathbf{a}_{_T}\},\,i=1,2,\ldots,N$ is a sequence of state-action pairs of the assembly of the belt insertion task, and N is the number of samples of demonstrations.

\section{Experimental Setup}
The considered assembly is the insertion of a rubber belt into a pulley, which is in WRS2020 (World Robot Summit). We build up the experimental simulation setup in Gazebo~9. We also use a real manipulator Universal Robot UR10e to obtain human correction demonstrations through compliance control at 500Hz based on a position/force hybrid controller. The learning algorithm is implemented with PyTorch \cite{pytorch}.

\section{Results}
\begin{figure}[h!]
\centering{
{
\includegraphics[width=245pt , height=245pt]{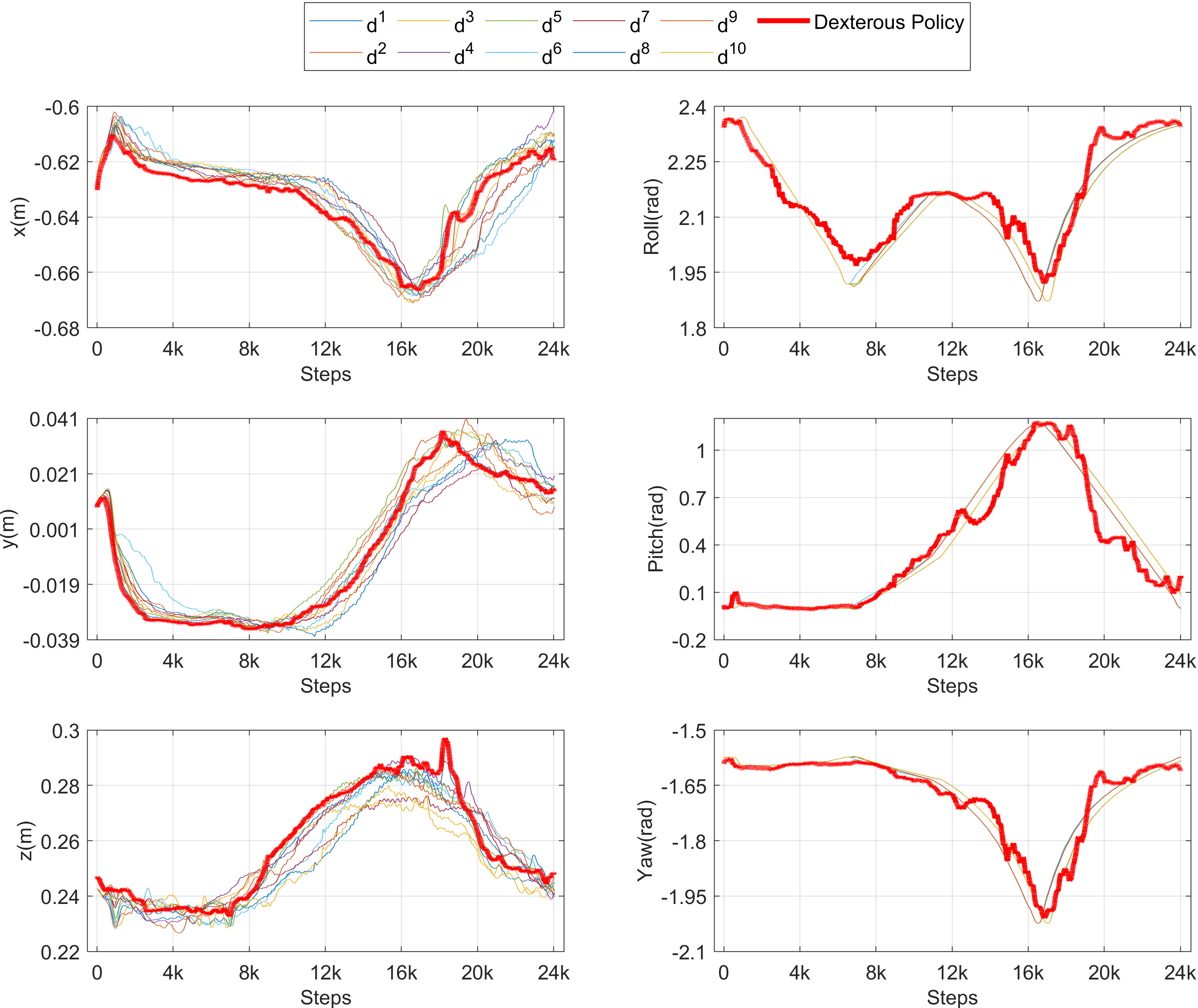}
}
\caption{
A learned path to finish the belt insertion task by combining 100 virtual demonstrations with 10 human ones. 
}\label{fig:learned_paths}
}
\end{figure}

 Fig.~\ref{fig:learned_paths} shows that a learned policy, dexterous policy generates reference paths (red line) for the given task based on demonstrations. In this work, we only correct positions of end-effector by a human because orientation trajectory from the offline paths is good enough to finish the given task. This means that it is not necessary to do a demonstration if the offline paths are fine. Lastly, we compared the learned path by the dexterous policy with human demonstrations $\{d^1,d^2,\ldots,d^{10}\}$, and the result presents that the combined approach is possible to learn a trajectory for a highly complex assembly task without not only a lot of human demonstrations starting from scratch but also external systems.
\section{Conclusion}
In this paper, we present an approach for generating paths to assemble a highly complex assembly task of deformable objects. Our approach contains three parts to obtain a final policy called dexterous policy. The first part is to generate offline paths from TrajOpt \cite{RN12}, and the second part is to provide the corrected paths by a human. In the third part, we use virtual human corrected paths. Since the offline paths are provided based on well-studied mathematical modeling of a deformable object, we have more reliability of the paths than a scratch to obtain them. Human corrections can also complement the discrepancy that the offline paths have. Moreover, in order to address the lack of data sets, we generate virtual human-guided data sets through the polynomial approximation approach. Therefore, we can have a lot of data sets despite doing a small number of human demonstrations. Finally, we obtain the dexterous policy using a neural network based on data sets. In conclusion, we generate reference paths to finish a highly complicated assembly task of deformable objects without any external systems like teleoperation and vision-based motion capture system. Our approach can be also used in highly complicated assembly tasks that human demonstrations are difficult to be taken place without external systems.
\label{sec:conclusion}

\section*{Acknowledgments}
This work was supported by MADE FAST.
\\
\\
\\
\bibliographystyle{unsrtnat}
\bibliography{references}

\end{document}